\documentclass{article} % For LaTeX2e
\usepackage{iclr2019_conference,times}

\usepackage[utf8]{inputenc} % allow utf-8 input
\usepackage{hyperref}       % hyperlinks
\usepackage{url}            % simple URL typesetting
\usepackage{amsfonts}       % blackboard math symbols
\usepackage{graphicx}       % figures
\usepackage{subcaption}    % subfigures
\usepackage{amsmath}  % equation references
\usepackage{paralist}
\usepackage[ruled, vlined, linesnumbered]{algorithm2e}
\newenvironment{algorithmicarchitecture}[1][htb]
  {% Update algorithm name
   \begin{algorithm}[#1]%
  }{\end{algorithm}}

\title{Autonomous Goal Exploration using Learned Goal Spaces for Visuomotor Skill Acquisition in Robots}

\author{
  Adrien Laversanne-Finot\\
  Flowers Team\\
  Inria and Ensta-ParisTech, France\\
  \texttt{adrien.laversanne-finot@inria.fr} \\
   \And
   Alexandre Péré \\
   Flowers Team \\
   Inria and Ensta-ParisTech, France\\
   \texttt{alexandre.pere@inria.fr} \\
   \AND
   Pierre-Yves Oudeyer \\
   Flowers Team \\
   Inria and Ensta-ParisTech, France\\
   \texttt{pierre-yves.oudeyer@inria.fr}
}

\iclrfinalcopy % Uncomment for camera-ready version, but NOT for submission.
\begin{document}

\maketitle

%===============================================================================

\begin{abstract}
%TODO A réécrire, au moins le début
%Intrinsically motivated goal exploration processes enable agents to autonomously sample goals to explore efficiently complex environments with high-dimensional continuous actions. They have been applied successfully to real world robots to discover repertoires of policies producing a wide diversity of effects. Often these algorithms relied on engineered goal spaces but it was recently shown that one can use deep representation learning algorithms to learn an adequate goal space in simple environments. Experiments have yet only been performed on simulated experiments. Here we presents recent results obtained on a real robotic setup where a 6-joints robotic arm evolves in an arena containing a ball.

% Alex: Quelques phrases qui peuvent être intégrées pour refaire l'intro peut être ? 
The automatic and efficient discovery of skills, without supervision, for long-living autonomous agents, remains a challenge of Artificial Intelligence. Intrinsically Motivated Goal Exploration Processes give learning agents a human-inspired mechanism to sequentially select goals to achieve. This approach gives a new perspective on the lifelong learning problem, with promising results on both simulated and real-world experiments. Until recently, those algorithms were restricted to domains with experimenter-knowledge, since the Goal Space used by the agents was built on engineered feature extractors. The recent advances of deep representation learning, enables new ways of designing those feature extractors, using directly the agent experience. Recent work has shown the potential of those methods on simple yet challenging simulated domains. In this paper, we present recent results showing the applicability of those principles on a real-world robotic setup, where a 6-joint robotic arm learns to manipulate a ball inside an arena, by choosing goals in a space learned from its past experience.
\end{abstract}

\section{Introduction}
Despite recent breakthroughs in artificial intelligence, learning agents often remain limited to tasks predefined by human engineers. The autonomous discovery and simultaneous learning of many tasks in an open world remains challenging for reinforcement learning algorithms. However, discovering autonomously the set of outcomes that can be produced by acting in an environment is of paramount importance for learning agents. This is essential to acquire world models and repertoires of parameterized skills \citep{Baranes2013, da2014active, hester2017intrinsically} or to efficiently bootstrap exploration for deep reinforcement learning problems with rare or deceptive rewards \citep{conti2017improving, colas18}. In order to discover as many diverse outcomes as possible, the learner should be able to self-organize its exploration curriculum in order to discover efficiently the possible outcomes that can be produced in its environment.

When aiming at discovering autonomously what outcomes can be produced by a physical robot, a naive exploration of the space of motor commands is bound to fail. Indeed, the space of motor commands is often continuous and high-dimensional. Secondly, this space is also highly redundant: many motor commands will produce the same effect. Lastly, in any real world setup, the number of samples that can be collected is limited. Thus, discovering diverse outcomes and learning policies to reproduce them requires more elaborate strategies.

One approach that was shown to be efficient in this context is known as Intrinsically Motivated Goal Exploration Processes (IMGEPs) \citep{baranes2010intrinsically, Forestier2017}, an architecture closely related to Goal Babbling \citep{Rolf2010}. The general idea of IMGEPs is to equip the agent with a goal space. During exploration, the agent will sample goals in this goal space according to a certain strategy, before trying to reach them using an associated goal-parameterized reward function. For each sampled goal the agent will dedicate a budget of experiments to improve his performance regarding this particular goal. Crucially, the agent stores each outcome discovered during the exploration, which allows him to learn in hindsight how to achieve each outcome he discovers, should he later sample it as a goal. This makes the approach powerful since targeting a goal will often allow an agent to simultaneously learn about other goals. IMGEPs can be implemented with population-based policy learning approaches \citep{baranes2010intrinsically, Pere2018} or using goal parameterized deep reinforcement learning techniques \citep{colas2018curious}.

Until recently IMGEPs where limited to engineered goal spaces. This approach limits the autonomy of the agent, and in many interesting problems, such a goal space is not provided and may be hard to design manually. It is always possible to use the sensory space as a goal space. However, in many cases the sensory space is high-dimensional, e.g. made of low perceptual measures such as pixels, and building a goal-parameterized reward directly in this space is problematic. Thus, it was proposed in \cite{Pere2018} to leverage representation learning algorithms such as Variational Auto-Encoders (VAEs) and to use the learned latent space as the goal space. It was further shown in \cite{laversanne-finot18a} that if the representation used as a goal space is disentangled (e.g. encoding separately different physical properties of the environment), then it becomes possible to achieve more efficient exploration in environments with multiple objects and distractors, through a modular goal exploration algorithm that samples goals which maximize the learning progress. 

However all these experiments were performed using simulated environments. Furthermore, they assumed the availability of many observations of outcomes produced by another agent, covering the diversity of possible outcomes, in order to train initially the goal space representation.

In this paper, we provide evidence that the ideas developed in those papers can also be successfully applied to real world scenarios. We also show how they can be transposed in a sample efficient manner to a fully autonomous learning setting where the representation learning mechanism is trained on outcomes data gathered autonomously by the agent. In particular we consider an experiment where a 6-joint robotic arm interacts with a ball inside a closed arena and we show that using a learned representation as a goal space leads to a better exploration of the environment than a strong baseline consisting in randomly sampling dynamic motion primitives.

%\footnote{The arm is controlled with dynamic motion primitives and is perceived through raw pixels.}

%The paper is organized as follows : first, the IMGEPs framework is introduced. Then, some results obtained on simulated environments are presented in order to illustrate the architectural principles of IMGEPs. Lastly we show our results on the robotic experiment.

\begin{figure}
\includegraphics[width=\textwidth]{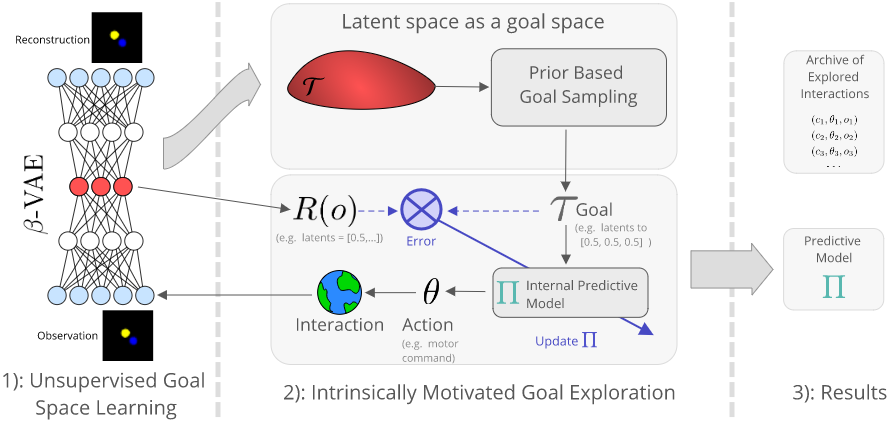}
\caption{The IMGEP with learned goal spaces strategy.}
\label{fig:imgep_ugl}
\end{figure}

\section{Goal exploration with learned goal spaces}

This section briefly introduces \emph{Intrinsically Motivated Goal Exploration Processes}, using a learned representation of the goal space. The overall architecture is summarized in \figurename~\ref{fig:imgep_ugl}. For a more thorough introduction to IMGEPs with engineered goal spaces and learned goal spaces we refer to \cite{Forestier2017} and \cite{laversanne-finot18a}, respectively.

In order to understand the general idea of IMGEPs, one must imagine the agent as performing a sequence of contextualized and parameterized experiments. At the beginning of each experiment the agent will in sequence: observe the context, sample a goal according to some strategy, use its internal knowledge (policy) to find the best motor parameters to achieve this goal in this context, and then perform the experiment using these parameters. The goals are arbitrary and can range from ``moving the ball to this specific position'' to ``moving the end effector of the arm  to this location'', when the goal space is hand-crafted. If this is not the case one strategy proposed in \cite{Pere2018} is to learn a representation of the environment, using data sampled from demonstrations, and to use the latent space as the goal space. In this case a goal is a point in the latent space, and one uses a similarity function in this space as the associated goal achievement reward function. The agent then tries to produce an outcome that, when encoded, is as close as possible to this point in the latent space. See Algorithmic Architecture~\ref{alg:imgep} for a high level algorithmic description of IMGEPs and Appendix~ \ref{ann:imgep} for more details on the different components.

%One can imagine more involved goal selection schemes \citep{laversanne-finot18a}. For example is it possible to define goals only on a subset of the latent variables. In this case a goal would be for example to perform an action that leads to an outcome whose encoding have the latent variable $1$ and $2$ as close as possible to $(0.2, 0.8)$. These goals selection schemes can prove useful when the representation is disentangled. In this case, each of the latent variables encode for a distinct factor of variation and so it is natural to see each (or a group) of them as distinct goals.

%In this description modules refer to different types of goals (e.g. ball position for the one module and arm position for the another module or different groups of latent variables when the goal space is a representation). See Appendix~\ref{ann:imgep} for more details on the module selection scheme. %TODO lister détails en appendice et change appendice en conséquence...

\begin{algorithmicarchitecture}
  \caption{Goal Exploration Strategy}
  \label{alg:imgep}
   \KwIn{\\
         Policy $\Pi$, History~$\mathcal{H}$, (optional) Goal space (engineered or learned): $(R, \gamma)$, }
   \Begin{
       \For{A fixed number of Bootstrapping iterations}{
         Append $\mathcal{H}$ using \textit{Random Motor Exploration}}
       Learn the goal space using a representation learning algorithm (if not provided) \\
       Initialize Policy $\Pi$ with history $\mathcal{H}$ \\
       \For{A fixed number of Exploration iterations}{
         Observe context $c$\\
         Sample a goal, $\tau \sim \gamma$ \\
         Compute $\theta$ using $\Pi$ on tuple $(c, \tau)$ \\
         Perform experiment and retrieve observation $o$ \\
         Append $(c, \theta, o)$ to $\mathcal{H}$ \\
         Update Policy $\Pi$ with $(c, \theta, o)$
         }
       }
  \KwRet{The history $\mathcal{H}$}
\end{algorithmicarchitecture}

\section{Experiments}

We carried out experiments on a real world environment to address the following questions:
\begin{compactitem}
%\item Does the quality of the representation impact the performances of the exploration algorithm? In other words, does an exploration algorithm with a better representation discovers more diverse outcomes on average?

% Alex: Je comprend pas ce que tu veux dire par il n'y a pas d'obvious choice pour le engineered goal space. Est ce que dire ca c'est pas se tirer une balle dans le pied? 
%\item How learned goal spaces compare to engineered goal spaces in terms of exploration performances? In particular how do they compare when there is no obvious choice for the engineered goal space?

%\item Is it possible to learn a representation that when used as a goal space for IMGEPs yields an exploration dynamics that is as efficient as an exploration dynamics obtained with an engineered goal space?

\item To what extent can the ideas developed in simulated environments be applied on a real world setup?

\item Does the dataset used to train the representation algorithm need to contain examples of all possible outcomes to learn a goal space that gives good performances during exploration? Can it be learned during exploration, as example of outcomes are collected? 
\end{compactitem}

\begin{figure}
\centering
\includegraphics[width=0.5\textwidth]{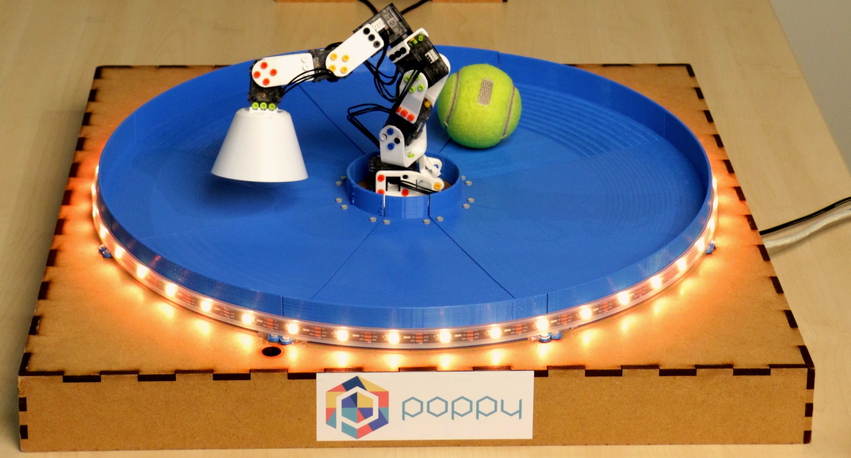}
\caption{The robotic setup. It consists of a 6-joint robotic arm and a ball that is constrained to move in an arena.}
\label{fig:ergoball}
\end{figure}

In order to answer those questions we experimented on a robotic setup that is similar in spirit to the environments considered in the simulated experiments and that we now describe in details:

\paragraph{Robotic environment} The environment is composed of a 6-joint robotic arm that evolves in an arena. In this arena a (tennis) ball can me moved around. Due to the geometry of the arena, the ball is more or less constrained to evolve on a circle. A picture of the environment is represented in \figurename~\ref{fig:ergoball}. The agent perceives the scene as a $64 \times 64$ pixels image. The motion of the arm is controlled by Dynamical Movement Primitives (DMP). Actions are the parameters of the DMPs used in the current episode. There is one DMP per joint. Each DMP is parametrized by one weight for each of the $7$ basis functions and one supplementary weight specifying the end joint state, for a total of $48$ parameters.

For the representation learning phase, we considered different strategies. In the first strategy (as was done in \cite{Pere2018} and \cite{laversanne-finot18a}) we consider that the agent has access to a database of examples of the possible set of outcomes. From this database the agent learns a representation that is then used as a goal space for the exploration phase. This strategy is referred to as \textbf{RGE (VAE)}. One could argue that using this method introduces knowledge on the set of possible outcomes that can be obtained by the agent. In order to test how this impacts the performances of the exploration algorithms we also experimented using a representation learned using only the samples collected during a the initial iterations of random motor exploration. We refer to this strategy as \textbf{RGE (Online)}.

%generated a set of images where the generative factors of the images (positions of the two balls in the simulated case and joints configuration of the robotic arm together with the ball position in the robotic case) were as uniformly distributed as possible. These images were then used to learn a representation using a VAE. 

\paragraph{Baselines}

Results obtained using IMGEPs with learned goal spaces are compared to two baselines:

\begin{itemize}
\item \textbf{Random Parameter Exploration (RPE)}, where exploration is performed by uniformly sampling parameters $\theta$. This strategy is inefficient as it does not leverage information collected during previous rollouts to choose the current parameters. It serves as a lower bound for the performances of the exploration algorithms. Since DMPs were designed to enable the production of a diversity of arm trajectories with only few parameters, this lower bound is already a reasonable baseline that performs better than applying random joint torques at each time-step of the episode.

\item \textbf{Goal Exploration with Engineered Features Representation (RGE-EFR)}: it is an IMGEP in which the goal space is handcrafted and corresponds (as closely as possible) to the true degrees of freedom of the environment. In this experiment it is not clear what is the best representation as multiple choices can be used (e.g. Cartesian or polar coordinates for the position of the ball). We settled for polar coordinates as the ball evolves on a circle. Since essentially all the information is available to the agent under a highly semantic form, it is expected to give an upper bound on the performances of the exploration algorithms.
\end{itemize}

\section{Results}
%TODO modify the number of trials
To assess the performances of the IMGEPs with learned goal spaces we performed between 8 and 14 trials for each of the configurations. In order to speed up the learning procedure, for each configuration using a learned goal space, we used the same representation for all trials\footnote{We did not pick a particular representation and preliminary experiments show that similar performances are obtained for other representations.}.

\begin{figure}
\centering
\includegraphics[width=0.75\textwidth]{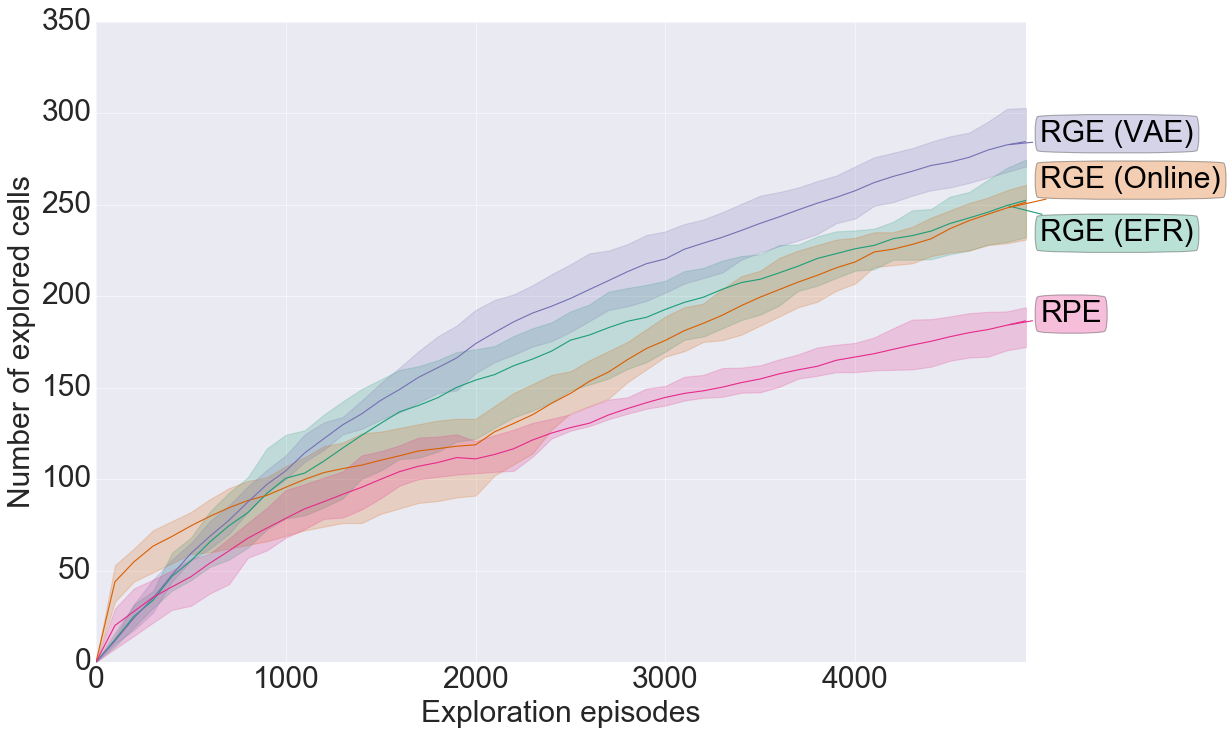}
\caption{Exploration performance during exploration.}
\label{fig:ExplorationPerfErgo}
\end{figure}

\paragraph{Exploration performances}

The performance of the algorithm is defined as the number of ball positions reached during the experiments. In this configuration, the ball is the hard part of the exploration problem since the end position of the robotic arm can be efficiently explored by performing random motor commands. In practice the performances of the exploration algorithms are measured by discretizing the outcome space in 900 cells (30 cells for each dimension) and counting the number of different cells reached by the ball during the experiment. The number of cells that can be reached is limited due to the finite size of the arm/arena.

The exploration performances are reported in \figurename~\ref{fig:ExplorationPerfErgo}. From the plot, it is clear that IMGEPs with both learned and engineered goal spaces perform better than the \textbf{RPE} strategy. When using a representation learned before exploration (\textbf{RGE (VAE)}) the performances are at least as good as exploration using the engineered representation. When the goal space is learned using the online strategy, there is an initial phase where the exploration performances are the same as \textbf{RPE}. However, after this initial collection phase, when the exploration strategy is switched from random parameter exploration to goal exploration using the learned goal space (at $2000$ exploration episodes) there is a clear change in the slope of the curve in favor of the goal exploration algorithm\footnote{Note that the first $2000$ exploration episodes for the online strategy are the same for all runs performed on the same platform. In practice it should be similar to the \textbf{RPE} curve which was performed with many more trials.}.

All in all, the differences in performances between IMGEPs and random parameter exploration are less pronounced than in past simulated experiments. We hypothesize that this is due to the ball being too simple to move around. Thus, the random parameter exploration, which leverages DMPs to produce diverse arm trajectories, achieves decent exploration results. Also the motors of the robotic arm are far from being as precise as in simulation, which makes it harder to learn a good inverse model for the policy and to output parameters that will move the ball.

\section{Conclusion}

In this paper we studied how learned representations can be used as goal spaces for exploration algorithms. We have shown in a real world experiment that using a representation as a goal space provides better exploration performances than a naive exploration of the space of motor commands.

One of the main advantages of using a learned goal space is that it alleviates the need to engineer a representation, which is not a simple task in general. For example, in the robotic setup it is not clear that the engineered representation used is the most convenient for the exploration algorithm. In this case, the position of the ball is parametrized using polar coordinates. In this representation two points that have the same distance to the center and have angles $0$ and $2\pi$ are perceived as very distant even though physically they correspond to the same outcome. Also the position of the ball is extracted using a handcrafted algorithm. It may happen that this algorithm fails (e.g. when the ball is hidden by the robotic arm). In that case it may report wrong values to the policy. Such problems make learning an inverse model harder and thus reduce the exploration performances. On the other hand, using a learned representation obliviates those problems.

%We have also shown that active exploration algorithms can leverage disentangled representations to discover independently controllable features. In the simulated experiment where there is a distractor the agent focus its exploration on controllable features which improves the exploration performances. On the real world experiment since there is no distractor the performances are similar. Yet the agent is able to discover that some features cannot be learned.

As mentioned in the paper, it is possible to imagine more involved goal selection schemes (see \ref{ann:modularimgep} for a short description of the results described in \cite{laversanne-finot18a}) when the representation is disentangled. These goal selection schemes leverage the disentanglement of the representation to provide better exploration performances. We tested these ideas in this experiment and did not find any advantages in using those goal selection schemes. This is not surprising since there are no distractors in this experiment and modular goal exploration processes are specifically designed to handle distractors. Consequently, designing a real-world experiment with distractors, in order to test modular goal exploration processes with learned goal spaces, would be of great interest for future work.

%===============================================================================

\clearpage
\subsubsection*{Acknowledgments}
We would like to thank Sébastien Forestier for help in setting up the experiment. Simulated experiments presented in this paper were carried out using the PlaFRIM experimental testbed, supported by Inria, CNRS (LABRI and IMB), Université de Bordeaux, Bordeaux INP and Conseil Régional d'Aquitaine (see \url{https://www.plafrim.fr/}).

%===============================================================================

\bibliography{papers}
\bibliographystyle{iclr2019_conference}

%===============================================================================

\clearpage

\section{Appendices}

\subsection{Intrinsically Motivated Goal Exploration Processes}\label{ann:imgep}

In this part, we give further explanations on Intrinsically Motivated Goal Exploration Processes. 

\paragraph{Meta-Policy Mechanism}\label{ann:metapolicy}

The (Meta-)Policy is responsible to outputs the actions/parameters that are used during the episode. Given a context $c$ and a goal $\tau$ the Policy should output the parameters $\theta$ that are the most likely to produce an observation $o$ that \emph{fulfills} the task $\tau$. That an observation $o$ \emph{fulfills} a task $\tau$ can be quantified by a cost function $C: \mathcal{T} \times\mathcal{O} \mapsto \mathbb{R}$.

\begin{figure}
\centering
\begin{subfigure}{.7\textwidth}
	\includegraphics[width=\textwidth]{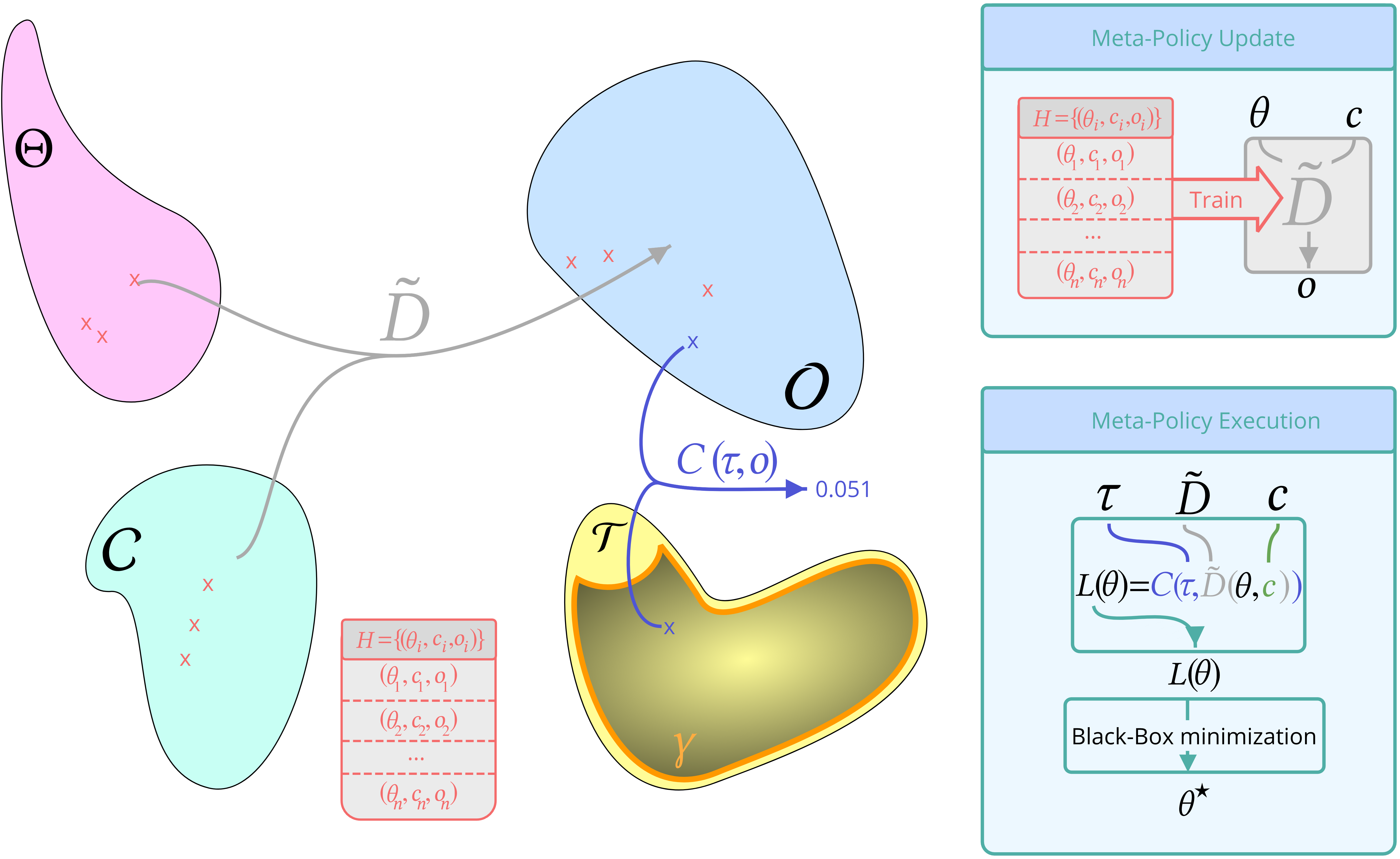}
	\caption{Direct-Model Meta-Policy}
\end{subfigure}
\begin{subfigure}{.7\textwidth}
	\includegraphics[width=\textwidth]{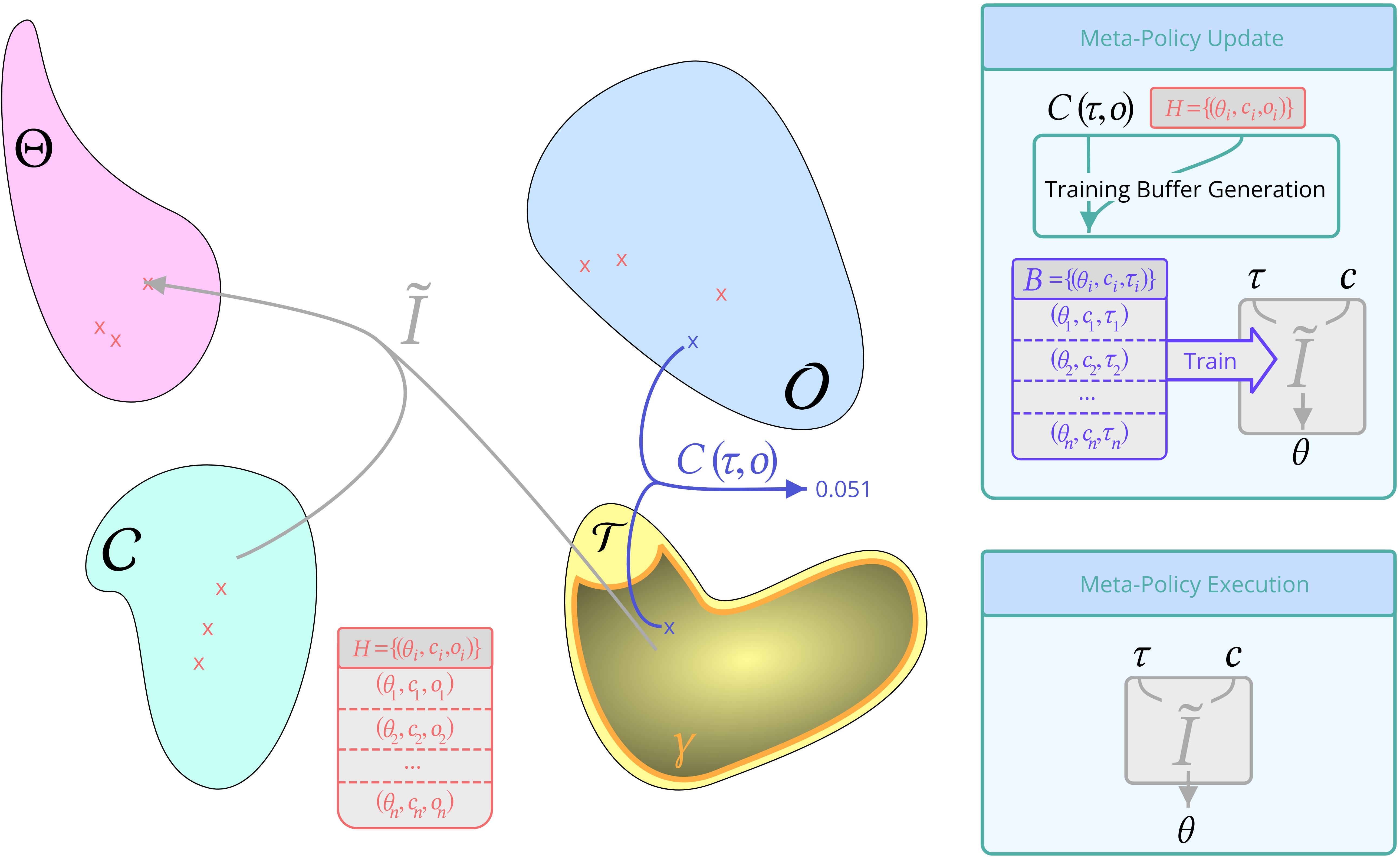}
	\caption{Inverse-Model Meta-Policy}
\end{subfigure}
\caption{The two different approaches to construct a meta-policy mechanism.}
\label{fig:meta-policies}
\end{figure}

There are two different ways to construct a meta-policy both which are depicted in \figurename~\ref{fig:meta-policies}:
\begin{compactitem}
\item \textbf{Direct-Model Meta-Policy:} In this case, an approximate phenomenon dynamic model $\tilde{D}$ is learned using a regressor (e.g. LWR). The model is then updated regularly by performing a training step with the newly acquired data. At execution time, for a given goal $\tau$, a loss function is defined over the parameterization space through $L(\theta)=C(\tau, \tilde{D}(\theta, c))$. A black-box optimization algorithm, such as L-BFGS, is then used to optimize this function and find the optimal set of parameters $\theta$ (see \citep{Baranes2013, Forestier2016, Benureau2016} for examples of such meta-policy implementations in the IMGEP framework). 

\item \textbf{Inverse-Model Meta-Policy:} In this approach, an inverse model $\tilde{I}: \mathcal{T} \times \mathcal{C} \mapsto \Theta$ is learned from the history $\mathcal{H}$ which contains all the previous experiments in the form of tuples $(c_i, \theta_i, o_i)$. To learn the inverse model it is necessary to associate to every observation $o_i$ a task $\tau_i$. The inverse model can then be learned using usual regression techniques from the set $\{( \tau_i, c_i, \theta_i )\}$.
\end{compactitem}

%In this case, we directly learn an inverse model $\tilde{I}:\mathcal{T}\times\mathcal{C}\mapsto\Theta$. To do that, a training buffer $\mathcal{B}$ containing elements $\{c_i, \tau_i, \theta_i\}$ must be generated out of the history $\mathcal{H}$. Put differently, for each sample of $\mathcal{H}$, we must find if a given goal is optimally solved by the outcome, and if it is append it to the buffer $\mathcal{B}$. Using this buffer a regressor can be trained as an inverse model.

In our case, we took the approach of using an Inverse-Model based Meta-Policy. We draw the attention of the reader on the following implementation details:
\begin{compactitem}
\item It may happen that using different parameters one obtain the same final outcome. For example different movements of the arm can put the ball and the arm in the same final position. However, in general, a combination of parameters leading to the them outcome does not produce a similar outcome. This is often referred to as the redundancy problem in robotics or as a multi-modality issue \citep{PathakICLR2018}. To tackle this issue, we used a $\kappa$-nn regressor with $\kappa=1$.

\item In order to associate to each of the observations a goal we used the (either learned or engineered) embedding function. To the observation $o_i$ corresponds the goal $\tau_i$ defined through: $\tau_i := R(o_i)$.
\end{compactitem}

Our particular implementation of the Meta-Policy is outlined in Algorithm~\ref{alg:train_meta}. The Meta-Policy is instantiated with one database per goal module. Each database store the representations of the observations projected on its associated subspace together with the associated contexts and parameterizations. Given that the meta policy is implemented with a nearest neighbor regressor, training the meta policy simply amounts to updating all the databases. Note that, as stated above, even though at each step the goal is sampled in only one module, the observation obtained after an exploration iteration is used to update all databases.

\begin{algorithm}[h]
  \caption{Meta-Policy (simple implementation using a nearest-neighbor model)}
  \label{alg:train_meta}
   \textbf{Require:} Goal modules: $\{R, P_k, \gamma(\tau|k), C_k \}_{k \in \{1, .., n_{mod} \} }$ \\
          \SetKwFunction{FMain}{Initialize\_Meta-Policy}
       \SetKwProg{Fn}{Function}{:}{}
       \Fn{\FMain{$\mathcal{H}$}}{
        \For{$k \in \{1, .., n_{mod} \}$}{
        $\text{database}_k \leftarrow \text{VoidDatabase}$ \\
            \For{$(c, \theta, o) \in \mathcal{H}$}{
               Add $(c, \theta, P_kR(o))$ to $\text{database}_k$
          }
      }
}
       \SetKwFunction{FMain}{Update\_Meta-Policy}
       \SetKwProg{Fn}{Function}{:}{}
       \Fn{\FMain{$c, \theta, o$}}{
        \For{$k \in \{1, .., n_{mod} \}$}{
               Add $(c, \theta, P_kR(o))$ to $\text{database}_k$
          }
}
\SetKwFunction{FMain}{Infer\_parameterization}
       \SetKwProg{Fn}{Function}{:}{}
       \Fn{\FMain{$c, \tau, k$}}{
   	       $\theta \leftarrow$ NearestNeighbor$(\text{database}_k, c, \tau)$ \\
           
          \KwRet{$\theta$}
}
\end{algorithm}

\subsection{Deep Representation Learning Algorithms}\label{ann:drl}

In this section we summarize the theoretical arguments behind Variational Auto-Encoder (VAE).

\paragraph{Variational Auto-Encoders (VAEs)} Let $\mathbf{x} \in \mathcal{X}$ be a set of observations. If we assume that the observed data are realizations of a random variable, we can hypothesize that they are conditioned by a random vector of independent factors $\mathbf{z}$, i.e. that $p(\mathbf{x} , \mathbf{z})= p(\mathbf{z}) p_{\theta}(\mathbf{x}, \mathbf{z})$, where $p(\mathbf{z})$ is a \textit{prior} distribution over $z$ and $p_{\theta}(\mathbf{x}, \mathbf{z})$ is a \textit{conditional distribution}. In this setting, given a i.i.d dataset $X = \{\mathbf{x}^1, \ldots, \mathbf{x}^N \}$, learning the model amount to searching the parameters $\theta$ that maximizes the dataset likelihood:
\begin{equation}
\log\mathfrak{L}(\mathcal{D}) = \sum_{i=1}^N \log p_{\theta} (\mathbf{x}^i)
\end{equation}
In practice it is often computationally intractable and so models are trained to optimize what is often referred to as the Evidence Lower Bound (ELBO):
\begin{equation}
\mathcal{L}(\mathbf{x}; \theta, \phi) = \mathbb{E}_{\mathbf{z}\sim q_{\phi}(\mathbf{z}|\mathbf{x})}[\log p_{\theta}(\mathbf{x}|\mathbf{z})] - \mathbb{D}_{KL}[q_{\phi}(\mathbf{z}|\mathbf{x}) \| p(\mathbf{z})],
    \label{eq:elbo}
\end{equation}
where $\mathbb{D}_{KL}$ is the Kullback-Leibler divergence, by jointly optimizing over the parameters (of often neural networks) $\theta$ and $\phi$.

\subsection{Details of Neural Architectures and training}\label{ann:detailsNN}

\paragraph{Model Architecture} The encoder for the VAEs consisted of 4 convolutional layers, each with 32 channels, 4x4 kernels, and a stride of 2. This was followed by 2 fully connected layers, each of 256 units. The latent distribution consisted of one fully connected layer of 20 units parametrizing the mean and log standard deviation of 10 Gaussian random variables. The decoder architecture was the transpose of the encoder, with the output parametrizing Bernoulli distributions over the pixels. ReLu were used as activation functions. This architecture is based on the one proposed in \cite{Higgins2016}.

\paragraph{Training details} The optimizer used was Adam \cite{Kingma2015}.

For the simulated experiment we used a learning rate of $5e^{-5}$ and batch size of 64. The overall training of the representation took 1M training iterations.

For the robotic experiment we used a learning rate of $1e^{-5}$ and batch size of $64$ and trained the network for 300k iterations when the representation was learned before the exploration. When the representation was learned with the outcomes obtained by the random exploration we used a batch size of $32$ the same learning rate and trained the network for $200k$ iterations.

\subsection{Scatter plots \textit{Robotic} environment}\label{ann:explo_ergo}

\begin{figure}
    \begin{subfigure}{.35\textwidth}
    \includegraphics[width=\textwidth]{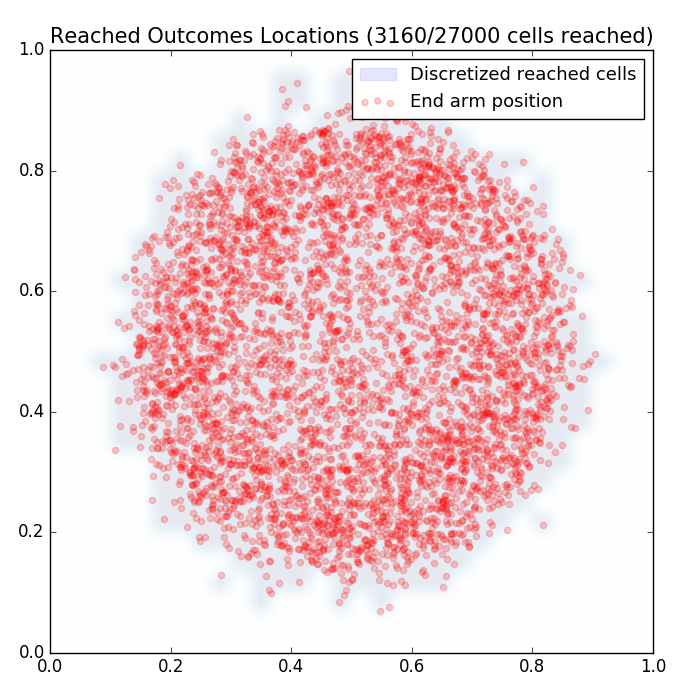}
    \caption{\textbf{RPE} - End of arm positions}
    \end{subfigure}
    \begin{subfigure}{.35\textwidth}
    \includegraphics[width=\textwidth]{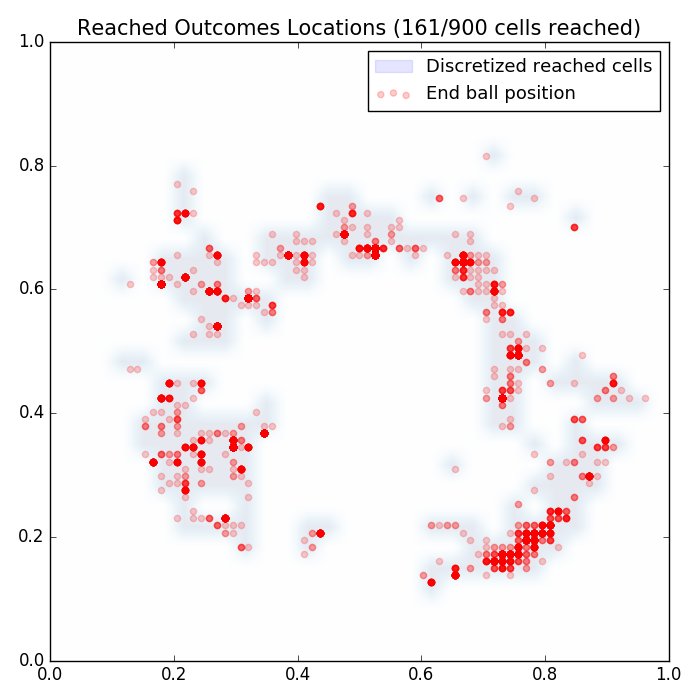}
    \caption{\textbf{RPE} - Ball positions}
    \end{subfigure}
    \\
    \begin{subfigure}{.35\textwidth}
    \includegraphics[width=\textwidth]{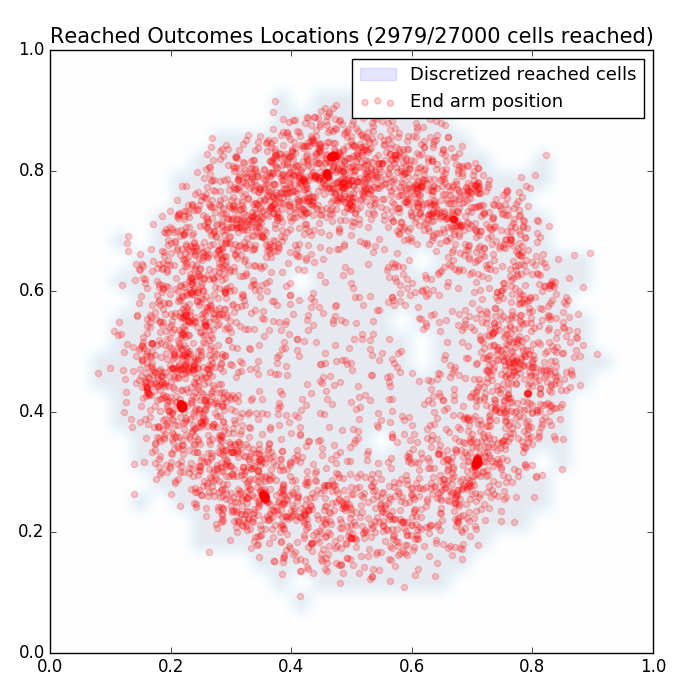}
    \caption{\textbf{RGE (EFR)} - End of arm positions}
    \end{subfigure}
    \begin{subfigure}{.35\textwidth}
    \includegraphics[width=\textwidth]{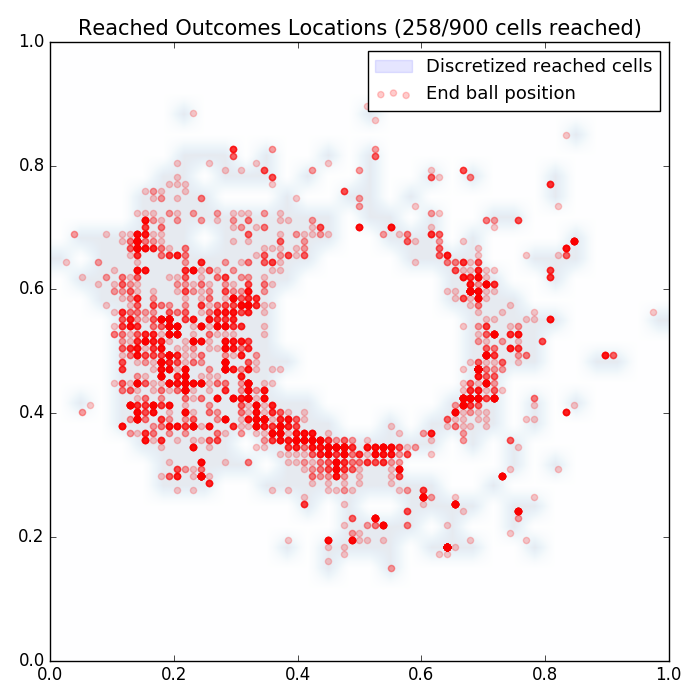}
    \caption{\textbf{RGE (EFR)} - Ball positions}
    \end{subfigure}
    \\
    \begin{subfigure}{.35\textwidth}
    \includegraphics[width=\textwidth]{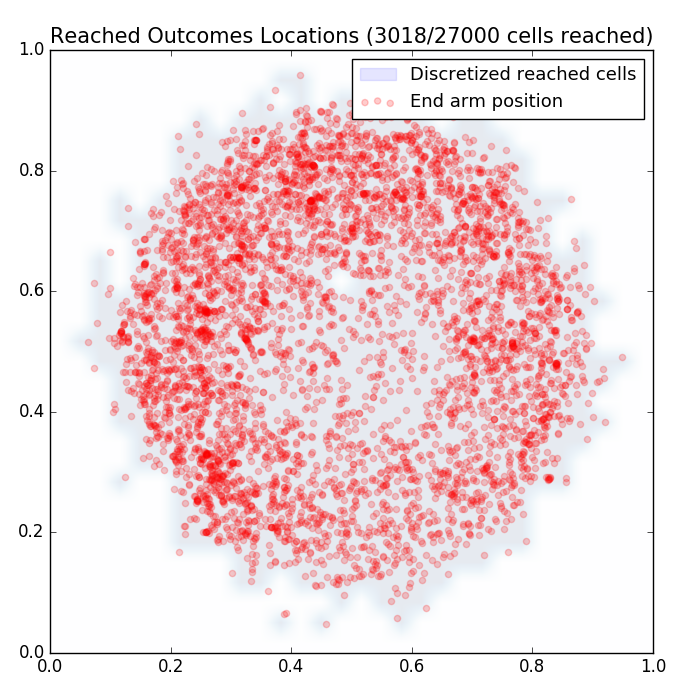}
    \caption{\textbf{RGE (VAE)} - End of arm positions}
    \end{subfigure}
    \begin{subfigure}{.35\textwidth}
    \includegraphics[width=\textwidth]{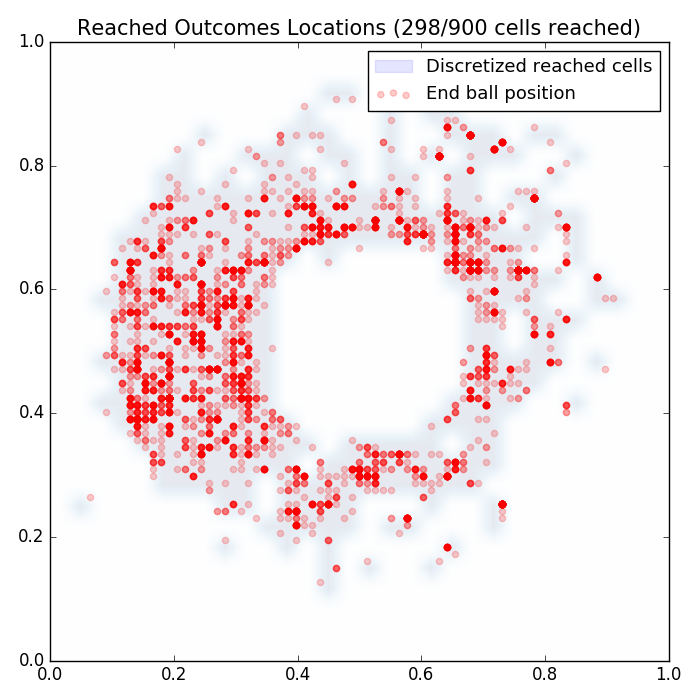}
    \caption{\textbf{MGE (VAE)} - Ball positions}
    \end{subfigure}
    \\
    \begin{subfigure}{.35\textwidth}
    \includegraphics[width=\textwidth]{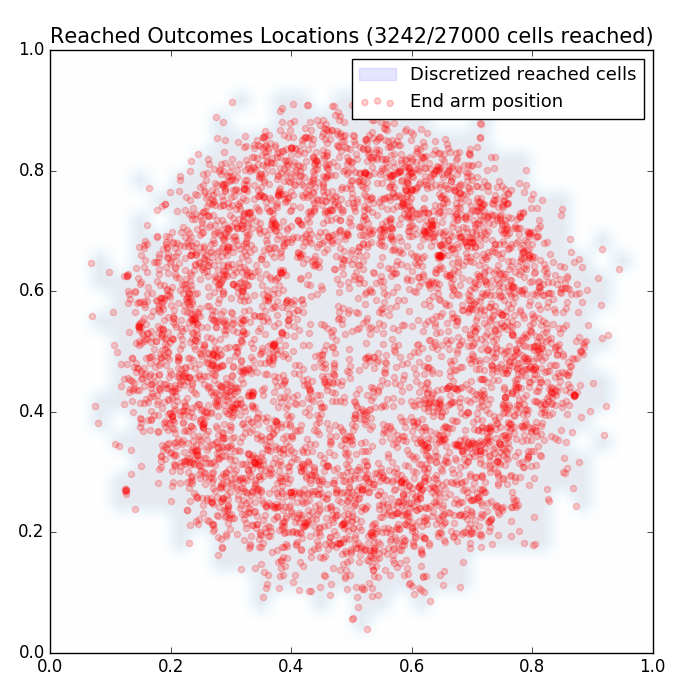}
    \caption{\textbf{Online} - End of arm positions}
    \end{subfigure}
    \begin{subfigure}{.35\textwidth}
    \includegraphics[width=\textwidth]{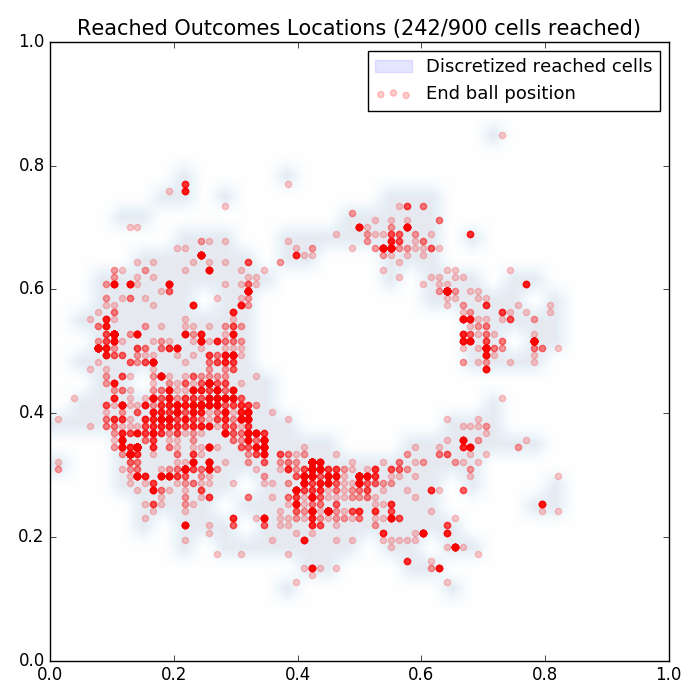}
    \caption{\textbf{Online} - Ball positions}
    \end{subfigure}
    \caption{Scatter plots of the end of arm and ball positions visited during exploration.}
    \label{fig:ScattersErgo}
\end{figure}

Scatter plots of the exploration for different exploration algorithms together with the number of cells reached are represented in \figurename~\ref{fig:ScattersErgo}. Although the exploration of the outcome space of the arm is similar for all algorithms there is a qualitative difference in the outcomes obtained in the outcome space of the ball between \textbf{RPE} and all instantiations of IMGEPs.

\subsection{Experimental setup}\label{ann:full_setup}

\begin{figure}
\centering
\includegraphics[width=.8\textwidth]{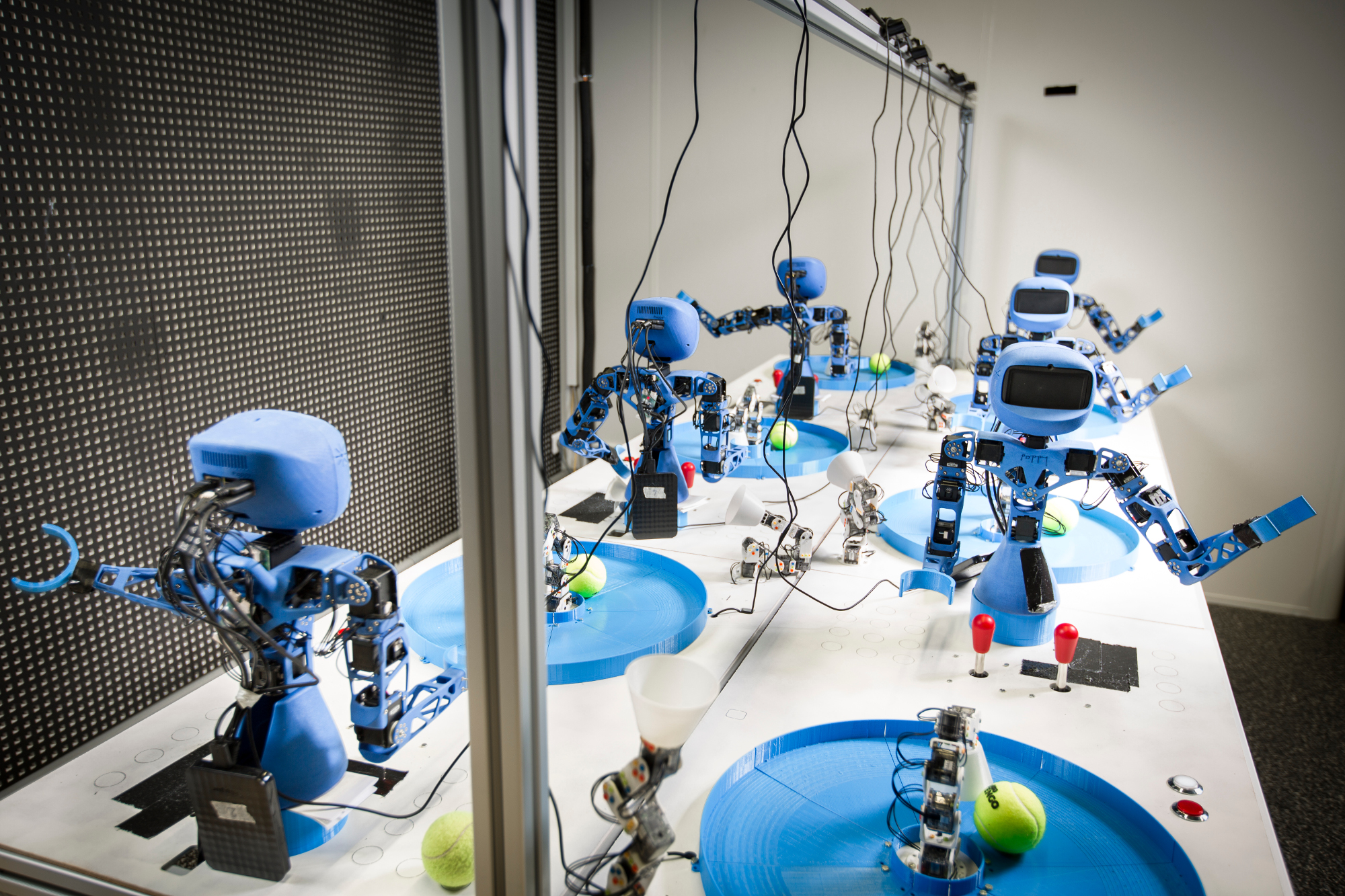}
\caption{Experiments are performed in parallel over 6 robots. In this experiment only the 6-joints robotic arm inside the arena is used.}
\label{fig:FullSetup}
\end{figure}

In practice experiments are performed in parallel using multiple copies of the same experiment. A picture of the complete experimental setup is represented in \figurename~\ref{fig:FullSetup}. Only the 6-joints robotic arm in the center of the arena is used in the experiments presented in this paper. Camera extracting the images are located on the bar above the setup.

\subsection{Modular Goal Exploration Processes}\label{ann:modularimgep}

In this section we recap some of the results presented in \cite{laversanne-finot18a}.

\subsubsection{IMGEPs with modular goal spaces}

As mentioned in the main text, when the environment is more complex and in particular when it contains distractors (objects that cannot be controlled), it is possible to design more efficient exploration algorithms. Modular goal exploration algorithms are designed to allow the agent to separate the exploration of different objects. For example the agent could decide to set for himself either goals for the ball or for its arm. The general idea is that some goals are harder (if not impossible) to reach than others. By monitoring its ability in fulfilling different kinds of goals the agent will be able to discover autonomously the difficulty of each type of goals and focus its exploration on goals which are neither too easy nor too hard. Using this strategy the agents thus autonomously design a curriculum. See Algorithmic Architecture~\ref{alg:modularimgep} for the corresponding algorithmic architecture.

When the goal space is engineered, the different modules can be readily defined when designing the goal space. However, in the case of learned goal spaces there is no easy solution. The strategy proposed in \cite{laversanne-finot18a} is to form \textit{modules} by grouping some of the latent variables together. The goals of one module are then to reach observations for which the latent variables corresponding to this module have specific values. If the representation of the world is disentangled, different latent variables encode for different degrees of freedom of the environment. In that case modules will correspond to distinct objects corresponding to the latent variables of this module. By monitoring its progress in controlling each of the latent variables the agent will discover that latent variables that encodes for distractors cannot be controlled while latent variables encoding for other objects can be controlled. The agent will thus be able to focus its exploration on controllable latent variables, leading to better exploration performances.

\begin{algorithmicarchitecture}
  \caption{Curiosity Driven Modular Goal Exploration Strategy}
  \label{alg:modularimgep}
   \KwIn{\\
         Goal modules (engineered or learned): $\{R, P_i, \gamma(\cdot| i), C_i\}$, Meta-Policy $\Pi$, History~$\mathcal{H}$}
   \Begin{
       \For{A fixed number of Bootstrapping iterations}{
         Observe context $c$\\
         Sample $\theta \sim \mathcal{U}(-1, 1)$\\
         Perform experiment and retrieve observation $o$\\
         Append $(c, \theta, o)$ to $\mathcal{H}$}
       Initialize Meta-Policy $\Pi$ with history $\mathcal{H}$ \\
       Initialize module sampling probability $p = \mathcal{U}(n_{mod})$ \\
       \For{A fixed number of Exploration iterations}{
         Observe context $c$\\
         Sample a module $i \sim p$\\
         Sample a goal for module $i$, $\tau \sim \gamma(\cdot| i)$ \\
         Compute $\theta$ using Meta-Policy $\Pi$ on tuple $(c, \tau, i)$ \\
         Perform experiment and retrieve observation $o$ \\
         Append $(c, \theta, o)$ to $\mathcal{H}$ \\
         Update Meta-Policy $\Pi$ with $(c, \theta, o)$ \\
         Update module sampling probability $p$ to follow learning progress
         }
       }
  \KwRet{The history $\mathcal{H}$}
\end{algorithmicarchitecture}

\subsubsection{Results on \textit{Arm-2-Balls}}\label{ann:armballs}

\begin{figure}
\includegraphics[width=\textwidth]{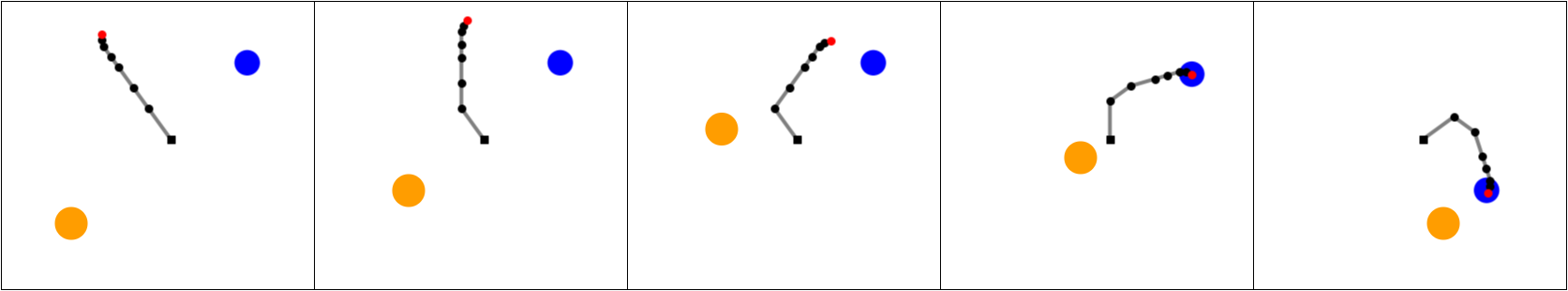}
\caption{A roll-out of experiment in the \textit{Arm-2-Balls} environment. The blue ball can be grasped and moved, while the orange one is a distractor that can not be handled, and follows a random walk.}
\label{fig:armballs}
\end{figure}

The ideas of modular IMGEPs were tested in the \textit{Arm-2-Balls} environment that is described below.

\textbf{Arm-2-Balls} The environment consists of a rotating 7-joint robotic arm that evolves in a scene containing two balls of different sizes, as represented in \figurename~\ref{fig:armballs}. One ball can be grasped and moved around in the scene by the robotic arm. The other ball acts as a distractor: it cannot be grasped nor moved by the robotic arm but follows a random walk. The agent perceives the scene as a $64 \times 64$ pixels image.

For the representation learning phase we used a Variational Auto-Encoder (VAE) for the entangled representation and a $\beta$-VAE for the disentangled representation. $\beta$-VAE are a variant of VAEs that have been argued to have better disentanglement properties \citep{Higgins2016, Higgins2017, Higgins2017a}. To train the representation, we generated a dataset of images for which the positions of the two balls were uniformly distributed over $[-1, 1]^4$. This dataset was then used to learn a representation using a VAE or a $\beta$-VAE. In order to test the impact of the disentanglement on the performances of the exploration algorithms, we used the same disentangled/entangled representation for all the instantiations of the exploration algorithms. This allowed us to study the effect of disentangled representations by eliminating the variance due to the inherent difficulty of learning such representations.

\begin{figure}
\begin{subfigure}{.5\textwidth}
  \centering
  \includegraphics[width=1.\linewidth]{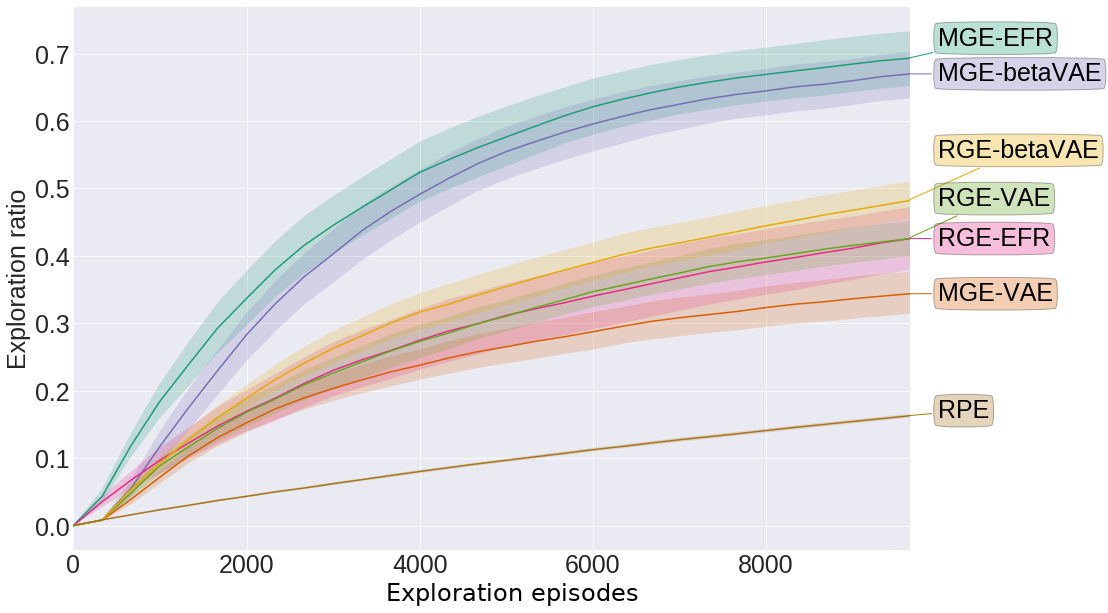}
  \caption{Small exploration noise ($\sigma = 0.05$)}
  \label{fig:ExplorationPerfSmallNoise}
\end{subfigure}
\begin{subfigure}{.5\textwidth}
  \centering
  \includegraphics[width=1.\linewidth]{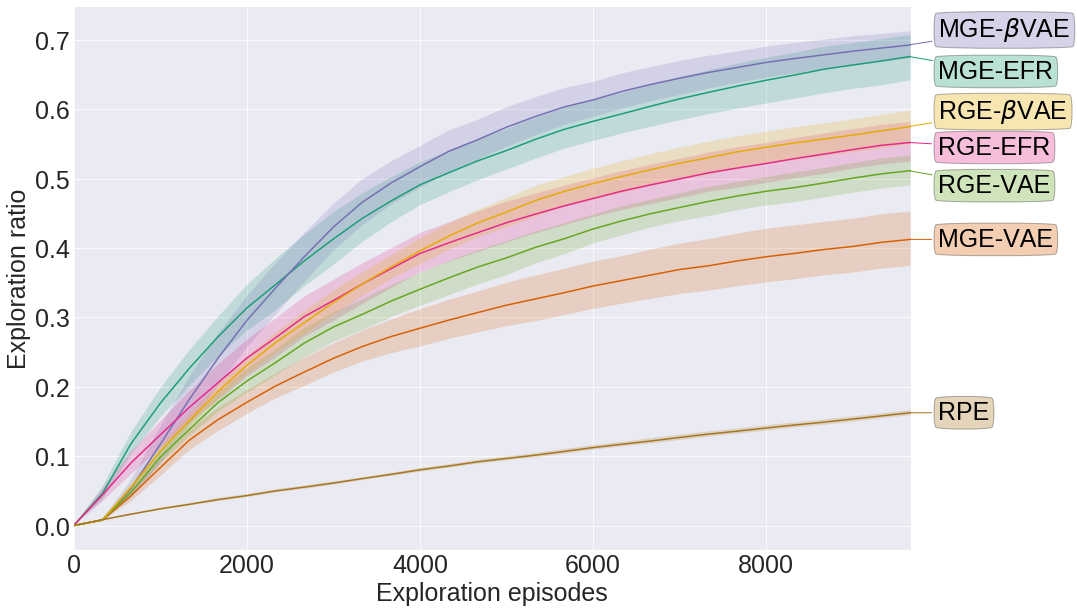}
  \caption{Large exploration noise ($\sigma = 0.1$)}
  \label{fig:ExplorationPerfLargeNoise}
\end{subfigure}%
\caption{Exploration ratio during exploration for different exploration noises.}
\label{fig:ExplorationPerfArmBalls}
\end{figure}

\subsubsection{Scatter plots \textit{Arm-2-Balls} environment}\label{ann:explo_armballs}

\begin{figure}
	\centering
    \begin{subfigure}{1.\textwidth}
    \includegraphics[width=\textwidth]{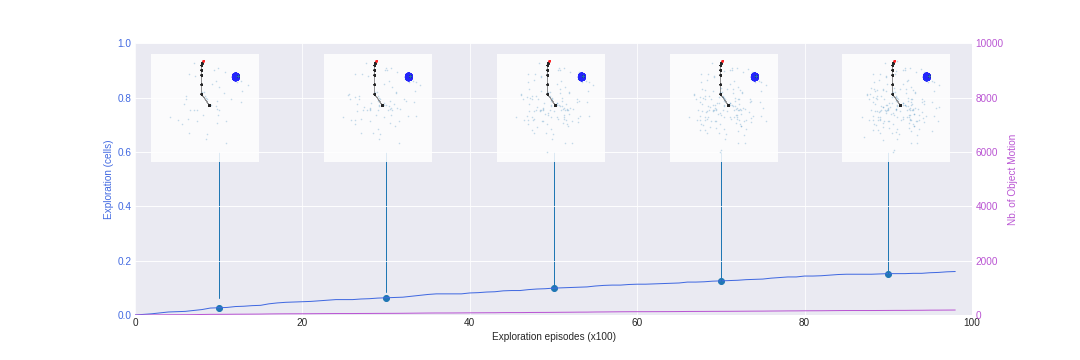}
    \caption{Random Parameterization Exploration}
    \end{subfigure}
    \begin{subfigure}{1.\textwidth}
    \includegraphics[width=\textwidth]{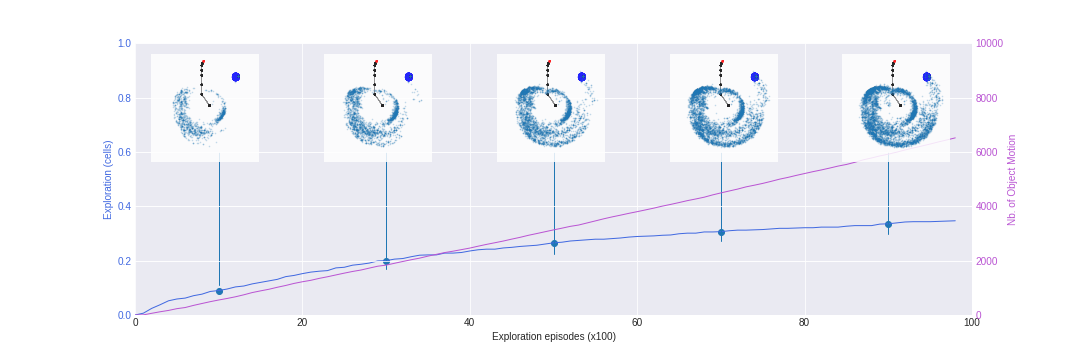}
    \caption{Random Goal Exploration with Engineered Features Representation (RGE-EFR)}
    \end{subfigure}
	\begin{subfigure}{1.\textwidth}
    \includegraphics[width=\textwidth]{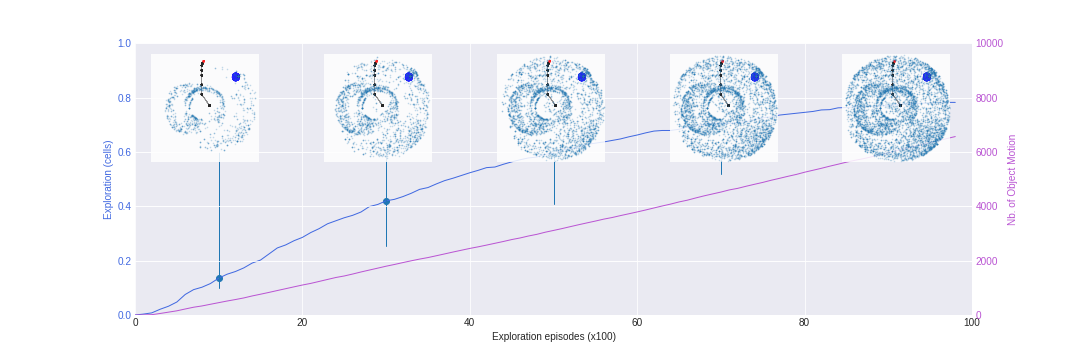}
    \caption{Modular Goal Exploration with Engineered Features Representation (MGE-EFR)}
    \end{subfigure}
    \caption{Examples of achieved observations together with the ratio of covered cells in the \textit{Arm-2-Balls} environment for \textbf{RPE}, \textbf{MGE-EFR} and \textbf{RGE-EFR} exploration algorithms. The number of times the ball was effectively handled is also represented.}
    \label{fig:explo_sprites}
\end{figure}

\begin{figure}
	\centering
    \begin{subfigure}{1.\textwidth}
    \includegraphics[width=\textwidth]{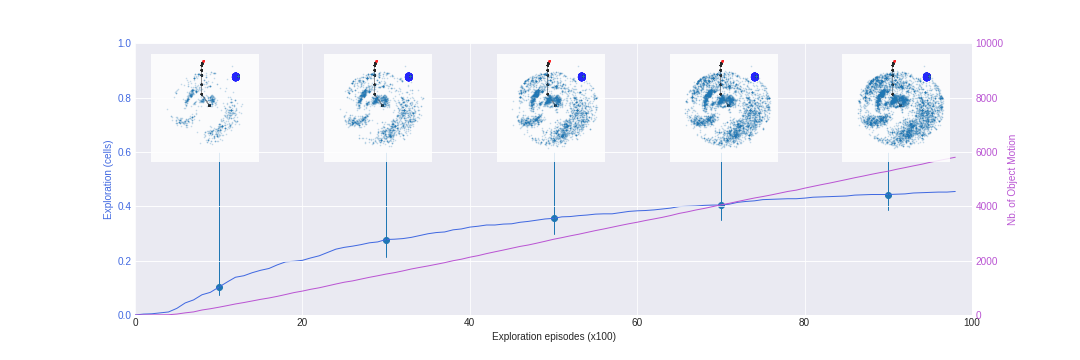}
    \caption{Random Goal Exploration with an entangled representation (VAE) as a goal space (RGE-VAE)}
    \end{subfigure}
	\begin{subfigure}{1.\textwidth}
    \includegraphics[width=\textwidth]{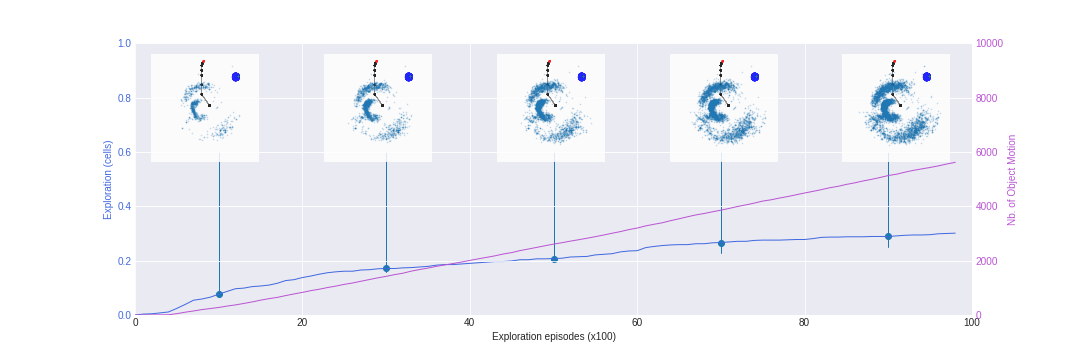}
    \caption{Modular Goal Exploration with an entangled representation (VAE) as a goal space (MGE-VAE)}
    \end{subfigure}
    \begin{subfigure}{1.\textwidth}
    \includegraphics[width=\textwidth]{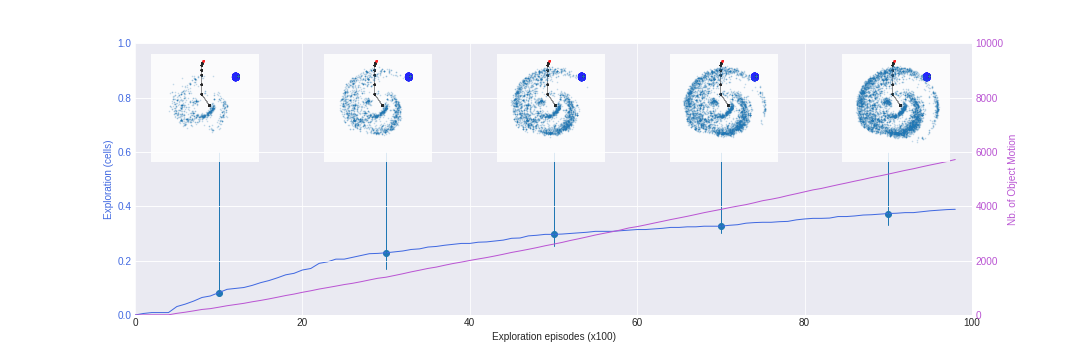}
    \caption{Random Goal Exploration with a disentangled representation ($\beta$VAE) as a goal space (RGE-$\beta$VAE)}
    \end{subfigure}
    \begin{subfigure}{1.\textwidth}
    \includegraphics[width=\textwidth]{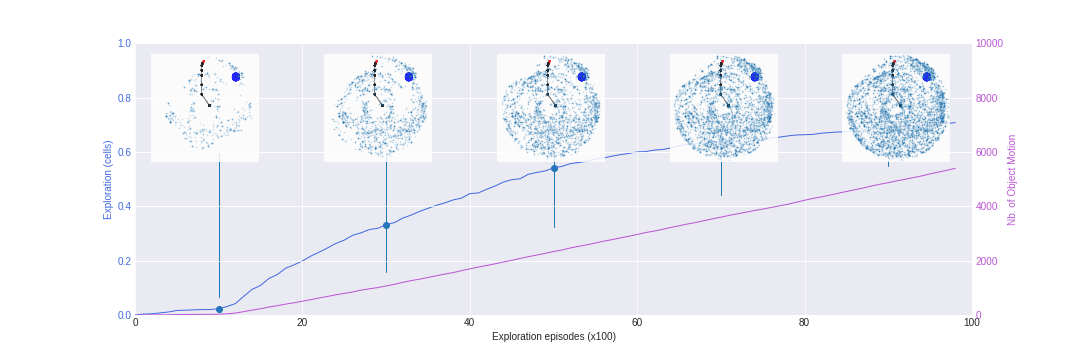}
    \caption{Modular Goal Exploration with a disentangled representation ($\beta$VAE) as a goal space (MGE-$\beta$VAE)}
    \end{subfigure}
    \caption{Examples of achieved observations together with the ratio of covered cells in the \textit{Arm-2-Balls} environment for \textbf{MGE} and \textbf{RGE} exploration algorithms using learned goal spaces (\textbf{VAE} and \textbf{$\beta$VAE}). The number of times the ball was effectively handled is also represented.}
    \label{fig:explo_sprites2}
\end{figure}

Examples of exploration curves obtained with all the exploration algorithms discussed in this paper (\figurename~\ref{fig:explo_sprites} for algorithms with engineered features representation and \figurename~\ref{fig:explo_sprites2} for algorithms with learned goal spaces). It is clear that the random parameterization exploration algorithm fails to produce a wide variety of observations. Although the random goal exploration algorithms perform much better than the random parameterization algorithm, they tend to produce observations that are cluttered in a small region of the space. On the other hand the observations obtained with modular goal exploration algorithms are scattered over all the accessible space, with the exception of the case where the goal space is entangled (VAE).

\end{document}